\documentclass{article}

\usepackage{arxiv}

\usepackage[utf8]{inputenc} 
\usepackage[T1]{fontenc}    
\usepackage{hyperref}       
\usepackage{url}            
\usepackage{booktabs}       
\usepackage{amsfonts}       
\usepackage{nicefrac}       
\usepackage{microtype}      
\usepackage{lipsum}
\usepackage{graphicx}

\usepackage{caption}
\usepackage{subcaption}

\graphicspath{ {./images/} }

\title{Time Series Generation with Masked Autoencoder}

\author{
 Mengyue Zha \\
  School of Mathematics\\
  The Hong Kong University of Science and Technology\\
  \texttt{mzha@ust.hk} \\
  \And
  SiuTim Wong\\
  School of Engineering\\
  The Hong Kong University of Science and Technology\\
  \texttt{stwongak@ust.hk} \\
   \And
  Mengqi Liu\\
  School of Statistics\\
  Beijing Normal University\\
  \texttt{mqliu@bnu.edu.cn} \\
  \And
 Tong Zhang \\
  School of Mathematics\\
  The Hong Kong University of Science and Technology\\
  \texttt{tongzhang@ust.hk} \\
  \And
 Kani Chen \\
  School of Mathematics\\
  The Hong Kong University of Science and Technology\\
  \texttt{makchen@ust.hk} \\
}

\begin{document}
\maketitle
\begin{abstract}
This paper shows that masked autoencoder with extrapolator (ExtraMAE) is a scalable self-supervised model for time series generation. ExtraMAE randomly masks some patches of the original time series and learns temporal dynamics by recovering the masked patches. Our approach has two core designs. First, ExtraMAE is self-supervised. Supervision allows ExtraMAE to effectively and efficiently capture the temporal dynamics of the original time series. Second, ExtraMAE proposes an extrapolator to disentangle two jobs of the decoder: recovering latent representations and mapping them back into the feature space. These unique designs enable ExtraMAE to consistently and significantly outperform state-of-the-art (SoTA) benchmarks in time series generation. The lightweight architecture also makes ExtraMAE fast and scalable. ExtraMAE shows outstanding behavior in various downstream tasks such as time series classification, prediction, and imputation. As a self-supervised generative model, ExtraMAE allows explicit management of the synthetic data. We hope this paper will usher in a new era of time series generation with self-supervised models. 
\end{abstract}


\section{Introduction}
Since embracing deep learning, time series models have witnessed a rapid growth in model capacity and capability. Latest models \cite{autoformer, informer, enhancing, nbeats} tend to have millions of trainable parameters. The explosion of model size relies on a sufficient supply of high-quality data. Many domains, however, fail to offer adequate qualified data. Most medical data contains sensitive personal information and thus cannot be shared directly. Financial data for events like flash crashes and great recessions are so scarce that studying the underlying mechanisms is nearly impossible \cite{jpmorgan}. Class imbalance and missing values may also corrupt data \cite{imputation_survey}. The appetite for data has now been the bottleneck of time series models in the era of deep learning. 

A natural idea to break the data limit is to generate synthetic data. Good synthetic data respects the temporal dynamics of original data and works as a substitute for the original data in real applications. One may use unsupervised Generative Adversarial Networks (GAN) \cite{gan} to generate synthetic data. GAN learns a mapping from random noises to the distribution of original data. Theoretically, a GAN can generate an arbitrary number of samples by sampling from random noises. A line of work has focused on GAN for time series generation. C-RNN-GAN \cite{cranngan} is the first attempt to generate synthetic time series by GAN. C-RNN-GAN implements generator and discriminator by Recurrent Neural Networks (RNNs). Later, RCGAN \cite{rcgan} upgrades C-RNN-GAN with additional condition information. In RCGAN, both generator and discriminator take conditions on auxiliary information. A multitude of GAN-variants generate synthetic data for specific domains such as medicine  \cite{scgan, nrgan}, finance \cite{cswgan}, sensor \cite{synsiggan} and decision making  \cite{datgan}. 

Learning temporal dynamics of original data lies at the heart of generating synthetic time series. Nevertheless, none of those above models consider the temporal nature of time series data. To better capture temporal dependencies in time series data, TimeGAN \cite{timegan} designs a supervised prediction task in the latent space. To our best knowledge, TimeGAN is the only time series generative model that successfully preserves the original data's temporal dynamics. Despite the success of supervised training in TimeGAN, unsupervised paradigms still dominate time series generation. We want to ask: what differs between unsupervised and supervised models for time series generation. We attempt to answer this question from the following perspectives: 
\begin{enumerate}
    \item The training difficulty differs. Supervised generative models seldom suffer from problems running in the family of unsupervised generative models. Some typical problems are non-convergence, vanishing gradient, and mode collapse \cite{stablizinggan}. Natural language processing (NLP) and Computer Vision (CV) address the problems through self-supervised pre-training. The solutions based on masking \cite{beit, bert, mae} are conceptually simple: removing a portion of data and learning to impute the removed content. 
    \item The maneuverability differs. Unsupervised generative models are born to be incompatible with extraneous manipulation. GAN takes random noises as seeds of generation and thus cannot systematically manage the generation. However, supervised generative models allow us to take fine control of generation. We may determine the diversity of synthetic data directly (See Supplementary Material). We may also generate synthetic data for specific downstream tasks such as classification, prediction, and imputation (See Section ~\ref{sec: experiments}). However, synthetic data generated by unsupervised models are blind to the downstream tasks. Unsupervised generative models cannot manage the diversity of synthetic data. They also fail to impute missing values due to a lack of supervision (See Section ~\ref{sec: experiments}). 
\end{enumerate}
The comparison provides us with sufficient reasons to abandon the unsupervised paradigm. We propose the masked autoencoder with extrapolator (ExtraMAE), a simple, effective, and scalable supervised model for time series generation. The training is self-supervised. In each iteration, ExtraMAE randomly masks some patches of the input time series and extrapolates the missing patches from unmasked patches. This design enables efficient and effective modeling of temporal dependencies. ExtraMAE proposes an extrapolator to reconstruct the missing patches in latent space. The extrapolator abandons mask tokens and thus avoids discontinuity in synthetic time series (See Supplementary Material). We also prune stacked Transformer blocks \cite{attention} which are pretty popular in previous masking-based models \cite{beit, bert, mae}. We replace the heavy Transformer blocks with lightweight RNNs. The pruning makes ExtraMAE the fastest model for time series generation in literature  (See Supplementary Material). The generation of ExtraMAE follows a cross-validation-like fashion (See Section ~\ref{sec: approach}). ExtraMAE scales well on different downstream tasks such as classification, prediction, and imputation (see Section ~\ref{sec: experiments}). It generates awesome synthetic time series even with an extreme mask ratio (e.g., $92\%$, see Section ~\ref{sec: experiments}). 

\section{Approach} 
\label{sec: approach}
\begin{figure}[htb]
	\begin{center}
	\includegraphics[width=0.98\textwidth]{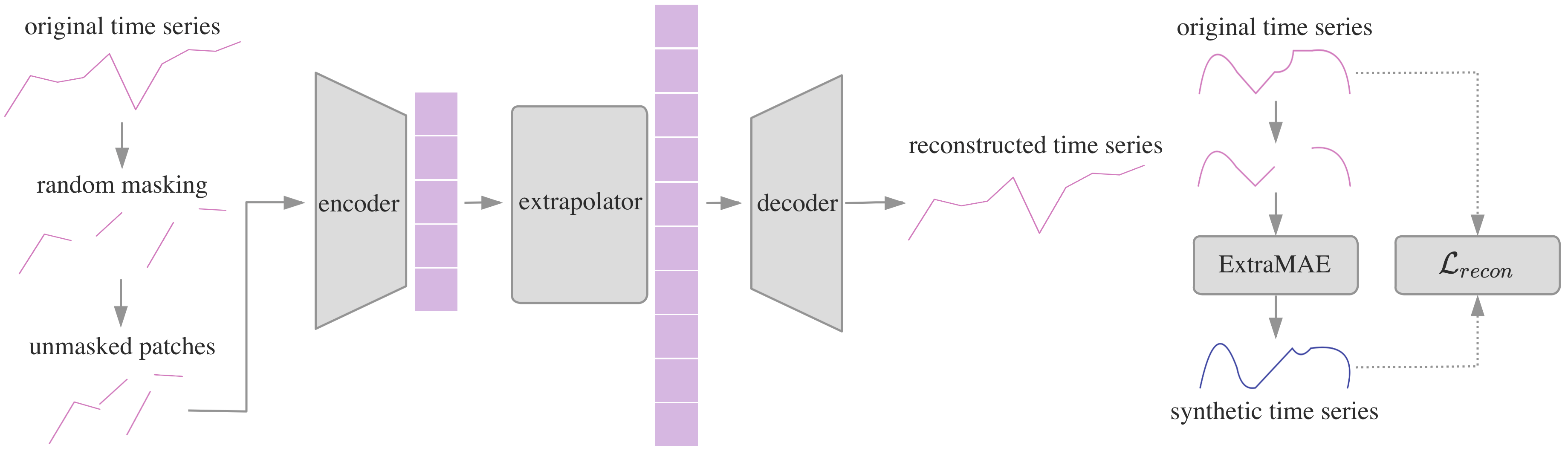}
	\end{center}
	\caption{Architecture (left) and training scheme (right) of ExtraMAE.}
	\label{fig: training}
\end{figure}

Masked Autoencoder with Extrapolator (ExtraMAE) recovers the original signal from its partial observations. ExtraMAE consists of three parts: an encoder, an extrapolator, and a decoder (See Figure ~\ref{fig: training}). The encoder only operates on unmasked patches. Once the masked patches are encoded, we extrapolate the latent representation of unmasked patches into latent representation for all patches (both masked and unmasked ones). Finally, the decoder maps the complete latent representation back into the feature space. We now present the problem statement and model architecture in detail. 

\textbf{Multivariate time series.} We denote a multivariate time series $X=[x_1|x_2|\cdots|x_L]\in \mathbb{R}^{d\times L}$ as a sequence of L observations. Each observation $x_i=(x_i^1, x_i^2, \cdots, x_i^d)^T \in\mathbb{R}^d$ consists of $d$ features where $i=1, 2, \cdots, L$. 

\textbf{Patches.} We slice a multivariate time series $X\in \mathbb{R}^{d\times L}$ into $T$ regular non-overlapping patches $\{P_1, P_2, \cdots, P_T\}$. The $j$-th patch $P_j\in \mathbb{R}^{d\times l}$ consist of $l=\frac{L}{T}$ consecutive observations $P_j=[x_{(j-1)l+1}, x_{(j-1)l+2}, \cdots, x_{jl+1}]$ where $j=1, 2, \cdots, T$. Here, both $T$ and $l$ are divisors of $L$. In Figure ~\ref{fig: training}, the original time series consists of nine line segments and thus we can slice it into $T=9$ patches. Each patch is a line segment. 

\textbf{Masking.} We sample a subset of $m$ patches to mask and reserve the remaining ones. The index set of masked patches $M=\{j_1, j_2, \cdots, j_m\}\subset \{1, 2, \cdots, T\}$.The index set of unmasked patches $N=\{1, 2, \cdots, T\}\setminus M=\{k_1, k_2, \cdots, k_n\}$ where $m+n=T$. For simplicity, we denote the invisible masked patches $[P_{j_1}|P_{j_2}| \cdots|P_{j_m}]$ as $P_{M}$ and visible unmasked patches $[P_{k_1}|P_{k_2}|\cdots|P_{k_n}]$ as $P_{N}$. Note that all elements in the masked patch $P_j\in\mathbb{R}^{d\times l}$ will be removed as a whole,  $j\in M=\{j_1, j_2, \cdots, j_m\}$. In Figure ~\ref{fig: training}, We randomly mask $m=4$ patches and concatenate the rest $n=5$ patches to get $P_N$. A line segment disappear as a whole once being masked. The index set of masked patches is $M=\{3, 5, 7, 9\}$. The index set of unmasked patches is $N=\{1, 2, 4, 6, 8\}$. 

\textbf{Encoder.} Our encoder only operates on visible, unmasked patches $P_{N}\in \mathbb{R}^{d\times (n\cdot l)}$. The encoder maps unmasked patches $P_N\in \mathbb{R}^{d\times (n\cdot l)}$ to its latent representation $H_{N}\in\mathbb{R}^{h\times(n\cdot l)}$. Here $h$ is the dimension of latent space. In Figure ~\ref{fig: training}, the encoder maps $P_N=[P_1|P_2|P_4|P_6|P_8]$ into their latent representations (purple squares on the left). We denote the encoder as a function $E:\prod_{k=1}^{n\cdot l} (\mathbb{R}^d)_k\rightarrow \prod_{k=1}^{n\cdot l}(\mathbb{R}^h)_k$ from unmasked patches $P_{N}$ to their latent representations $H_{N}$. $$H_{N} = E(P_{N}).$$ We implement $E$ as a stacked RNN followed by a fully connected layer. Note that the encoder only operates on visible patches and no mask tokens are used. 

\textbf{Extrapolator.} The extrapolator recovers the latent representations of masked positions. We denote the extrapolator as a function $I:\prod_{k=1}^{n\times l}(\mathbb{R}^h)_k\rightarrow \prod_{k=1}^{T\times l}(\mathbb{R}^h)_k$ that extrapolate the missing latent representations from visible neighbours. In Figure ~\ref{fig: training}, the extrapolator infers latent representations $\tilde{H}$ for all patches (purple squares on the right) from latent representations $H_N$ for unmasked patches (purple squares on the left). We implement $I$ as a fully connected layer. $$\tilde{H} = I(H_N)$$ 

\textbf{Decoder.} The decoder reconstructs original signals from the extrapolated complete latent representations. We denote the decoder as a function $D:\prod_{k=1}^{T\cdot l}(\mathbb{R}^{h})_k\rightarrow \prod_{k=1}^{T\cdot l} (\mathbb{R}^d)_k$ that maps latent representations $\tilde{H}\in\mathbb{R}^{h\times(T\cdot l)}$ into synthetic time series $\hat{X}\in\mathbb{R}^{d\times(T\cdot l)}$. Same as the encoder, we implement the decoder $D$ as a stacked RNN followed by a fully connected layer. $$\hat{X} = D(\tilde{H})$$ 

\begin{figure}[htb]
	\begin{center}
		\includegraphics[width=0.6\textwidth]{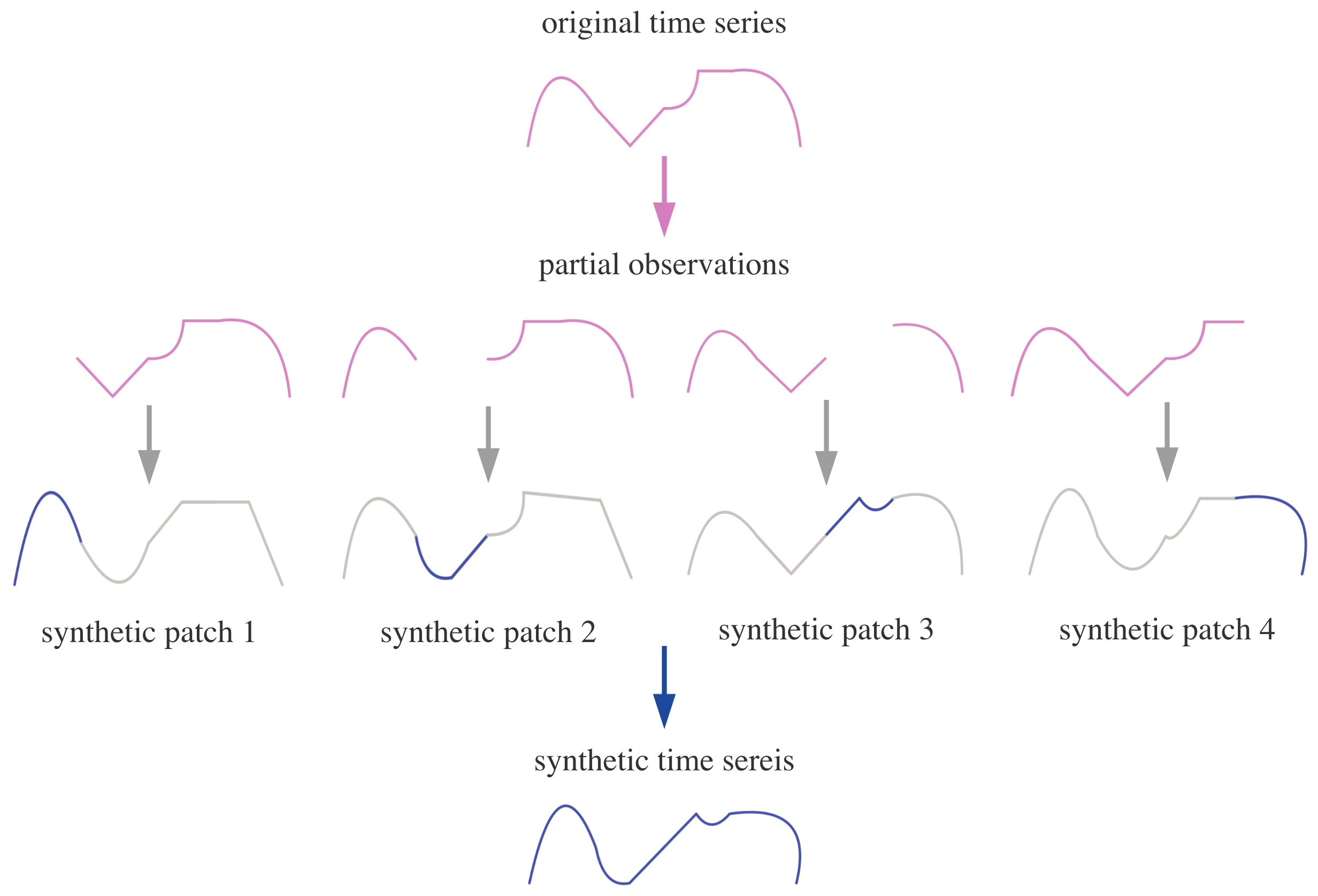}
	\end{center}
	\caption{Generation scheme of ExtraMAE. In this example, we divide the original time series into $4$ folds. ExtraMAE masks one fold each time and reconstructs the whole time series from the partial observation. We reserve the synthetic folds (blue line) and concatenate them into a synthetic time series. }
	\label{fig: generation}
\end{figure}

\textbf{Training.} We first randomly mask some patches of the original time series $X$. Then, ExtraMAE takes in unmasked patches $P_N$ and generates the reconstructed time series $\hat{X}$. Our loss function computes the mean squared error (MSE) between reconstructed time series $\hat{X}$ and original time series $X$ in $\mathbb{R}^{d\times(T\cdot l)}$. The expected reconstruction loss is $$\mathcal{L}_{recon}= \mathbb{E}_{X}||X-\hat{X}||_2.$$ Unlike \cite{bert, mae} which only computes loss on masked patches, we compute loss on all patches. The holistic reconstruction loss $\mathcal{L}_{recon}$ ensures that the reconstructed time series $\hat{X}$ preserves the continuity of the original time series $X$. 

\textbf{Generation} 
ExtraMAE generates synthetic data in a cross-validation like fashion (See Figure ~\ref{fig: generation}). We divide an original time series $X$ into several folds. Each time ExtraMAE removes one fold and reconstructs it from the partial observation. We repeat this process until every fold has a synthetic twin. We concatenate synthetic twins in order and then get the complete synthetic time series. Surprisingly, ExtraMAE generates superb synthetic time series even under extremely mask ratios. We also try masking several folds simultaneously. See Section ~\ref{sec: experiments} and the Supplementary Material for details. 

\textbf{Avoid mask tokens.} In \cite{beit, bert, mae}, a decoder plays two roles: (i) recovering information for missing positions. (ii) mapping the latent representations back into feature space. This design requires mask tokens. Mask tokens are special tokens that indicate the presence of masked patches. Decoders in previous studies take incomplete latent representation $H_N\in \mathbb{R}^{h\times (n\cdot l)}$ as input and hence need mask tokens as the indicators of missing patches. Mask tokens work well in models for NLP \cite{beit, bert} but they do not fit in time series generation. In contrast with the highly discrete latent space of text data, the latent space of time series data tends to be highly continuous. Mask tokens result in synthetic time series that are too discrete to preserve the continuity of the original time series (See Supplementary Material). Therefore, we avoid mask tokens in ExtraMAE. ExtraMAE proposes an extrapolator to recover the latent representations $H_M\in \mathbb{R}^{h\times (m\cdot l)}$ of masked patches. The decoder then maps the complete latent representation $\tilde{H}\in \mathbb{R}^{h\times (T\cdot l)}$ back into feature space. Since the decoder takes complete latent representation $\tilde{H}\in \mathbb{R}^{h\times (T\cdot l)}$ as the input, we do not need mask tokens anymore. We also avoid mask tokens in the encoder by only operating on visible unmasked patches $P_N$. 

\textbf{Statistical interpretation} Masked autoencoder learns temporal dynamics of original data by recovering the original signal from its partial observations. Statistical imputation \cite{little} regards this process as extrapolation density estimation. Extrapolation density $p(P_M|P_N)$ is the conditional probability of invisible masked patches $P_M$ given visible, unmasked patches $P_N$. Since extrapolation density contains direct information on temporal dependencies, ExtraMAE captures the temporal dynamics effectively and efficiently. An example of extrapolation density estimation is prediction. Predictors estimate extrapolation density on monotone missing data. In other words, prediction imputes the unknown future conditioned on the known past. However, monotone data is just a subset of non-monotone data \cite{little}. Extrapolation density estimated on non-monotone missing values is more informative than the one estimated on monotone data. ExtraMAE estimates extrapolation density on more general non-monotone data. This grounds for ExtreMAE beats TimeGAN. 

Before comparing ExtraMAE with other benchmarks, we ask: what is an excellent generative model? There are three desiderata for a time series generative model. (1) fidelity: the produced synthetic data should preserve the temporal dynamics of the original data. (2) practicality: the produced synthetic data should be able to substitute the original training data in real applications. (3) maneuverability: the generative model should allow users to manage the synthetic data explicitly. For example, a user may want to generate time series for a specific task (e.g., prediction, classification, and imputation) or take explicit control of the fidelity of synthetic data. This section shows that ExtraMAE outperforms all benchmarks in these three aspects. 

\section{Experiments}
\label{sec: experiments}

\begin{figure}[hbt]
     \centering
     \begin{subfigure}[b]{0.15\textwidth}
         \centering
         \includegraphics[width=\textwidth]{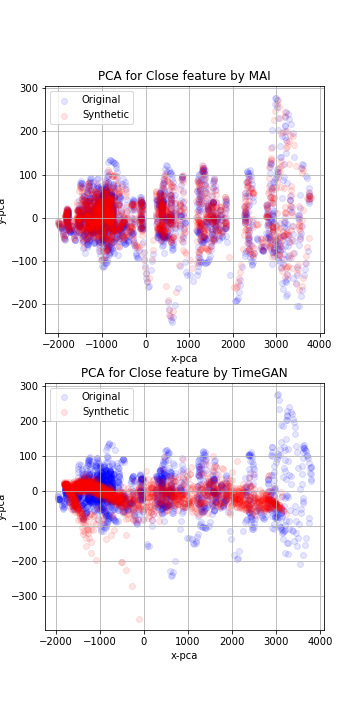}
         \caption{Close}
     \end{subfigure}
\hfill
     \begin{subfigure}[b]{0.15\textwidth}
         \centering
         \includegraphics[width=\textwidth]{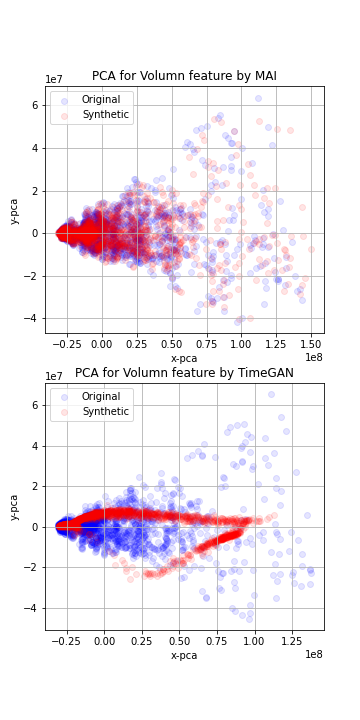}
         \caption{Volumn}

     \end{subfigure}
\hfill
     \begin{subfigure}[b]{0.15\textwidth}
         \centering
         \includegraphics[width=\textwidth]{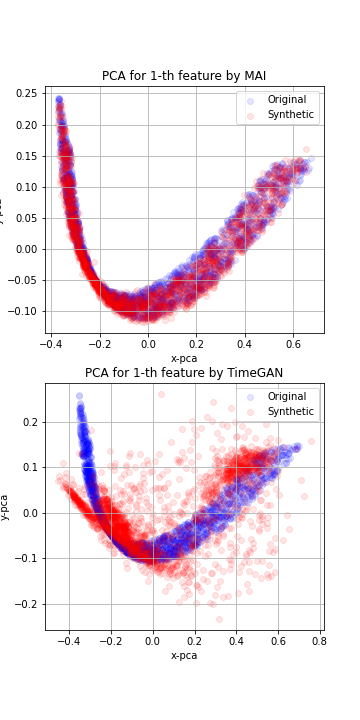}
         \caption{Sine$1$}
     \end{subfigure}
\hfill
     \begin{subfigure}[b]{0.15\textwidth}
         \centering
         \includegraphics[width=\textwidth]{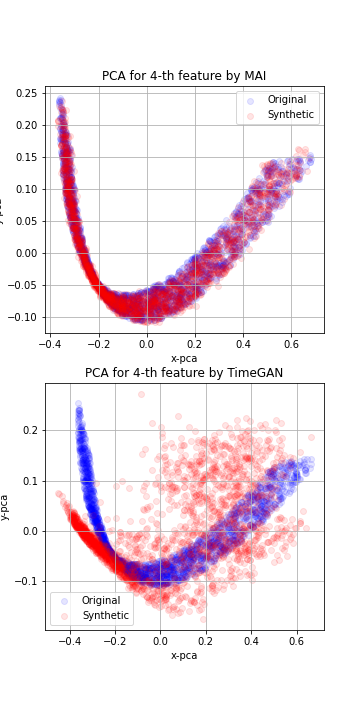}
         \caption{Sine$4$}
     \end{subfigure}
\hfill
     \begin{subfigure}[b]{0.15\textwidth}
         \centering
         \includegraphics[width=\textwidth]{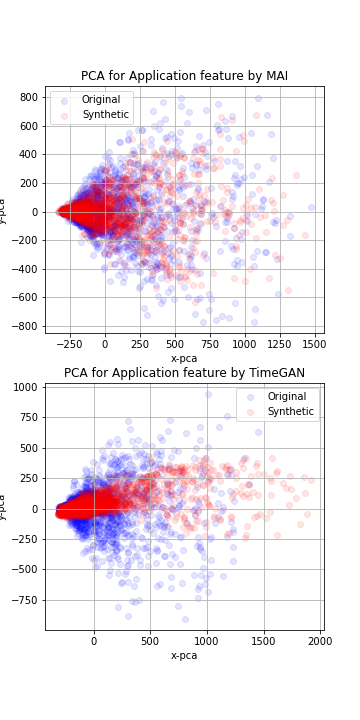}
         \caption{Application}
     \end{subfigure}
\hfill
     \begin{subfigure}[b]{0.15\textwidth}
         \centering
         \includegraphics[width=\textwidth]{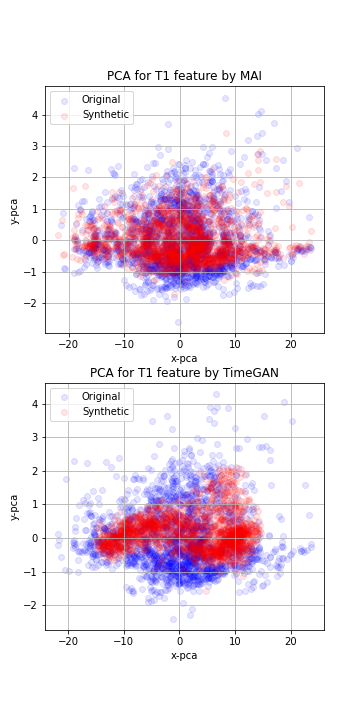}
         \caption{T$1$}
     \end{subfigure}
        \caption{PCA visualization of ExtraMAE (1st row) and current SoTA TimeGAN (2nd row) on six randomly selected features ((a)-(f)). Each column provides the visualization of one feature. }
        \label{fig: pca}
\end{figure}

\begin{figure}[hbt]
     \centering
     \begin{subfigure}[b]{0.15\textwidth}
         \centering
         \includegraphics[width=\textwidth]{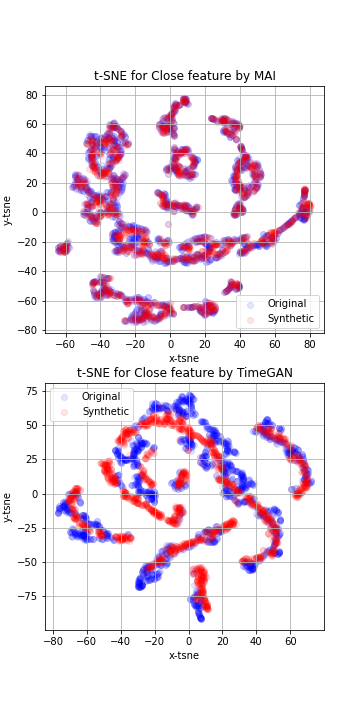}
         \caption{Close}
     \end{subfigure}
\hfill
     \begin{subfigure}[b]{0.15\textwidth}
         \centering
         \includegraphics[width=\textwidth]{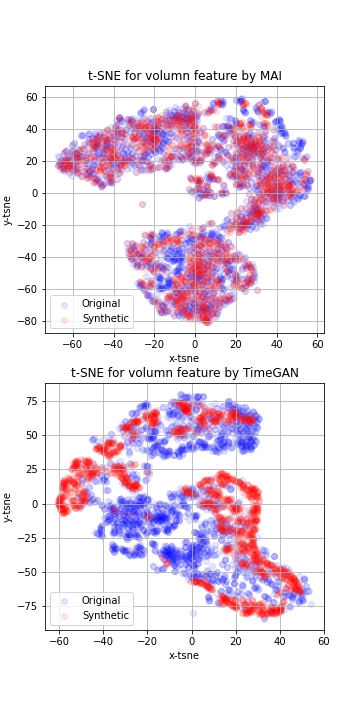}
         \caption{Volumn}

     \end{subfigure}
\hfill
     \begin{subfigure}[b]{0.15\textwidth}
         \centering
         \includegraphics[width=\textwidth]{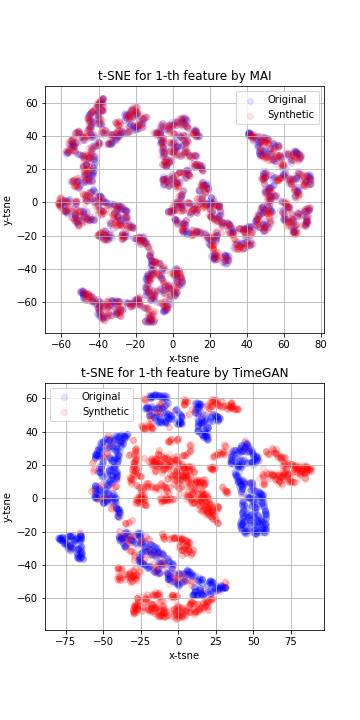}
         \caption{Sine$1$}
     \end{subfigure}
\hfill
     \begin{subfigure}[b]{0.15\textwidth}
         \centering
         \includegraphics[width=\textwidth]{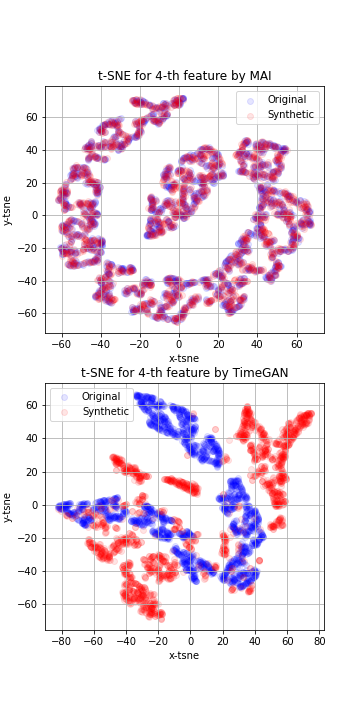}
         \caption{Sine$4$}
     \end{subfigure}
\hfill
     \begin{subfigure}[b]{0.15\textwidth}
         \centering
         \includegraphics[width=\textwidth]{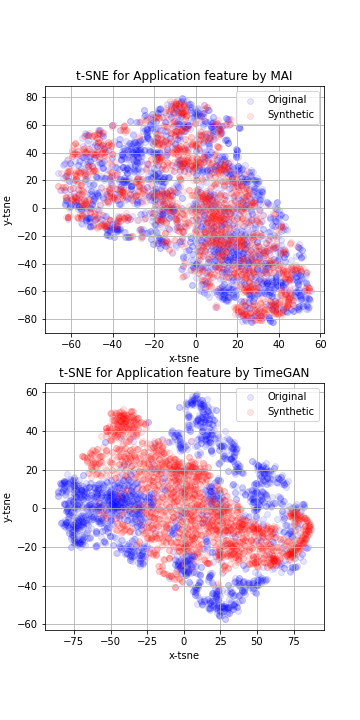}
         \caption{Application}
     \end{subfigure}
\hfill
     \begin{subfigure}[b]{0.15\textwidth}
         \centering
         \includegraphics[width=\textwidth]{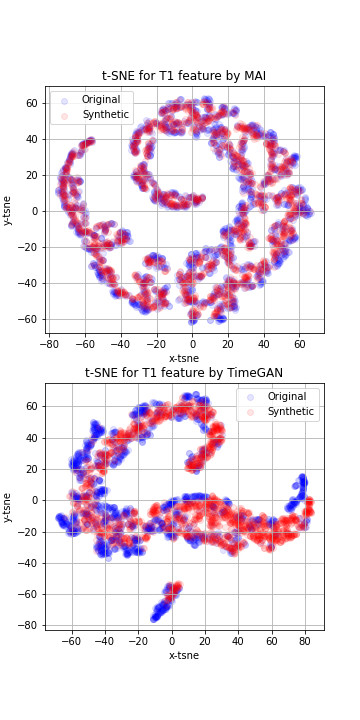}
         \caption{T$1$}
     \end{subfigure}
        \caption{t-SNE visualization of ExtraMAE (1st row) and current SoTA TimeGAN (2nd row) on six randomly selected features ((a)-(f)). Each column provides the visualization of one feature. }
        \label{fig: tsne}
\end{figure}

\subsection{Visualization}
First, we visualize original and synthetic data to evaluate fidelity intuitively. We apply t-SNE \cite{tsne} and PCA \cite{pca} to show how closely the distribution of the synthetic data resembles the distribution of the original data. Faithful synthetic data share similar distributions with the original data. 

\textbf{Baseline: TimeGAN.} In visualization, we compare ExtraMAE with TimeGAN. TimeGAN is the only benchmark that gives acceptable results in visualization. It is also the current state-of-the-art method for time series generation. TimeGAN adds a supervised prediction loss to the unsupervised GAN and learns temporal dynamics through the prediction loss. See Supplementary Material for visualizations of other benchmarks (i.e., RCGAN and C-RNN-GAN). 

\textbf{Visualization} We visualize TimeGAN-generated data $vs.$ ExtraMAE-generated data by PCA (figure ~\ref{fig: pca}) and t-SNE (figure ~\ref{fig: tsne}), respectively. We project the synthetic data (red dots) and the original data (blue dots) into a 2D plane and see how well the red and blue dots overlap. Faithful synthetic data should overlap with the original data. In Figures ~\ref{fig: pca} and ~\ref{fig: tsne}, we compare ExtraMAE (1st row) with the current SoTA TimeGAN (2nd row) on six features. We randomly pick up these six features from three datasets: Stocks, Sines, and Energy. We will discuss these three datasets later. ExtraMAE shows strikingly better fidelity than TimeGAN in both PCA plots (figure ~\ref{fig: pca}) and t-SNE plots (figure ~\ref{fig: tsne}). Synthetic data (red dots in the 1st row) generated by ExtraMAE is in perfect sync with the original data (blue dots). In contrast, the synthetic data generated by TimeGAN (red dots in the 2nd row) are in different manifolds from the original data (blue dots). Visualizations of other benchmarks are in Supplementary Materials. 

subsection{Discrimination scores} 
We then quantitatively evaluate the fidelity by discrimination scores. We train a post hoc discriminator (a 2-layer LSTM \cite{lstm}) to distinguish synthetic data from original data \cite{sctimegan}. We denote the classification accuracy of the post hoc discriminator as discrimination scores. The discriminator gives low discrimination scores when spurious synthetic data fools it. Therefore, low discrimination scores mean faithful synthetic data. Note that there is a trade-off between fidelity and diversity for synthetic data. We first show that ExtraMAE satisfies fidelity and then demonstrate its advantage in balancing fidelity and diversity (See Supplementary Material). We evaluate our model on various datasets with different statistics to get more reliable results. The datasets differ in periodicity, discreteness, regularity, level of noise, and degree of correlation. We discuss model hyperparameters and statistics of these three datasets in the Supplementary Material.   

\textbf{Stock.} The stock dataset \cite{stock} records daily historical Google stock data from 2004 to 2019. The stock dataset has five features: high, low, opening, closing, adjusted closing prices, and volume. Aperiodic and strong correlations between features characterize this continuous-value dataset. 

\textbf{Sine.} The sine dataset is simulated from sinusoidal random functions. Each sample has five independent features. Each feature is an univariate sinusoidal random function $x_i(t) = {\rm sin}(2\pi ft+\phi)$ where $i\in \{1,2,3,4,5\}$, frequency $f\sim {\rm U}(0,1)$ and phase $\phi \sim {\rm U}(-\pi, \pi)$. $\rm U$ stands for uniform distribution. The sine dataset is characterized by continuity, periodicity, and independence between features. 

\textbf{Energy.} The energy dataset \cite{energy} stands for UCI Appliances Energy Prediction dataset. This dataset records the energy consumption of appliances in a building. The energy dataset is continuity, noisy periodicity, and high dimensionality. 

\begin{table}[ht]
\caption{Discrimination scores}
\label{tab:disc_scores}
\begin{center}
\begin{tabular}{llll}
\toprule
Method & Stock & Sine & Energy \\
\midrule
ExtraMAE & $.071\pm .020$ & $.019 \pm .010$ & $.410 \pm .107$\\
TimeGAN & $.256\pm .036$ & $.184 \pm .071$ & $.499 \pm .001$\\
RCGAN & $.488 \pm .001$ & $.451 \pm .033$ & $.499 \pm .000$\\
C-RNN-GAN & $.500\pm .001$ & $.500\pm .000$ & $.499 \pm .000$\\
\cmidrule(lr){1-4}
Original & $.009\pm .005$ & $.009\pm .004$ & $.014\pm .003$\\
\bottomrule
\end{tabular}
\end{center}
\end{table}

Table ~\ref{tab:disc_scores} summarizes the discrimination scores of all candidate models on three datasets. ExtraMAE produces the most indistinguishable synthetic data. The discrimination scores of ExtraMAE are $72\%$, $90\%$, $18\%$ lower than the runner-up in Stock, Sine and Energy respectively ($72\% \approx \frac{|0.071-0.256|}{0.256}$, $90\%\approx\frac{|0.019-0.184|}{0.184}$, $18\%\approx\frac{|0.410-0.499|}{0.499}$). We also calculate the optimal discrimination scores in theory. The theoretical optimal discrimination score is achieved when the synthetic data is an exact copy of the original data. We mark the optimal discrimination scores in row \textit{Original} under the subtable for discrimination scores (See Table \ref{tab:disc_scores}). To make a more comprehensive comparison, we introduce RCGAN and C-RNN-GAN as benchmarks in addition to TimeGAN. 

\textbf{RCGAN.} RCGAN \cite{rcgan} is a conditional GAN model focusing on medical time-series data. RCGAN takes additional information from associated labels and uses stacked LSTM layers to process the condition recurrently. 

\textbf{C-RNN-GAN.} C-RNN-GAN \cite{cranngan} is a GAN model for continuous sequential data. It implements the generator by unidirectional RNN \cite{rnn} but uses bidirectional RNN \cite{birnn} for the discriminator. C-RNN-GAN also freezes the discriminator when its loss is $70\%$ less than the loss of the generator to prevent it from being too strong. 

\subsection{TSTR experiments}
We then show the practicality of ExtraMAE. Practical synthetic data should be able to substitute for original data in real applications. Therefore, we evaluate the practicality of synthetic data under \textit{Train on Synthetic Test on Real} (TSTR) framework \cite{rcgan}. In TSTR, we train a model by synthetic data but test it on the original data. If the synthetic data is practical, the model trained by synthetic data is supposed to perform similarly to the model trained by original data (See Supplementary Material). We evaluate the practicality of ExtraMAE and benchmarks by TSTR on two supervised tasks: prediction and classification. Table ~\ref{tab:pred_scores} and Table ~\ref{tab:class_scores} show how well the synthetic data work under TSTR on prediction and classification. Synthetic data generated by ExtraMAE consistently act as the best substitute for the original data across all datasets. To make a fair comparison, we follow the implementation of predictor and classifier in TimeGAN \cite{sctimegan}. 

\textbf{Prediction scores} We consider training a predictor on synthetic data. If synthetic data are excellent substitutes for original data, the predictor should give accurate predictions on original test data. We train the predictor by optimizing a 2-layer LSTM on the synthetic time series \cite{sctimegan}. The prediction score is the predictor's mean absolute test error on the original time series. Lower prediction scores indicate more practical synthetic time series. Table \ref{tab:pred_scores} lists the prediction scores of all candidate generative models on Sine, Stock, and Energy. We note that ExtraMAE consistently achieves the best performance on TSTR prediction. Prediction scores of ExtraMAE are $30.2\%$, $18.5\%$, $15.2\%$ lower than the runner-ups in Stock, Sine and Energy respectively ($30.2\%\approx \frac{|0.037-0.053|}{0.053}$, $18.5\%\approx\frac{|0.101-0.124|}{0.124}$, $15.2\%\approx\frac{|0.256-0.302|}{0.302}$). The theoretical infimum of the TSTR prediction score is achieved when the original data train the predictor. We mark the theoretical infimum in row \textit{Original} in Table \ref{tab:pred_scores}. 

\begin{table}[htb]
    \caption{Prediction scores}
    \label{tab:pred_scores}
    
        \begin{center}
        \begin{tabular}{llll}
            \toprule
            Method & Stock & Sine & Energy \\
            \midrule
            ExtraMAE & $.037\pm .000$ & $.101 \pm .000$ & $.256 \pm .001$\\
            TimeGAN & $.053\pm .001$ & $.124 \pm .001$ & $.302 \pm .002$\\
            RCGAN & $.375\pm .013$ & $.291 \pm .000$ & $.303 \pm .001$\\
            C-RNN-GAN & $.086 \pm .001$ & $.749 \pm .001$ & $.498 \pm .000$\\
            \cmidrule(lr){1-4}
            Original & $.036\pm.001$ & $.102\pm .001$ & $.249\pm .000$\\
            \bottomrule
        \end{tabular}
        \end{center}
\end{table}

\textbf{Classification scores} We consider training a classifier on synthetic data. The classifier maps unlabeled time series to a set of labels. We implement the classifier as a 3-layer multilayer perceptron (MLP) \cite{mlp}. Since practical synthetic data are supposed to preserve temporal dynamics necessary for classification, the classifier is supposed to make an accurate classification of original data. We denote the test accuracy of the classifier as the classification score. Large classification scores indicate practical synthetic data. Table \ref{tab:class_scores} lists the classification scores of all candidate generative models on three labeled time-series datasets: Wafer \cite{wafer}, IPD \cite{ipd}, and Strawberry \cite{berry}. We note that ExtraMAE consistently achieves the highest classification scores. Classification scores of ExtraMAE are $9.9\%$, $32.5\%$, $9.4\%$ higher than the runner-ups in Wafer, IPD and Strawberry respectively ($9.9\%\approx \frac{|72.3-65.8|}{65.8}$, $32.5\%\approx\frac{|83.9-63.3|}{63.3}$, $9.4\%\approx\frac{|61.4-56.1|}{56.1}$). The theoretical maximum of the TSTR prediction score is achieved when the original data train the classifier. In Table \ref{tab:class_scores}, we mark the the classification scores of classifier trained by original data in row \textit{Original}. Note that synthetic data for classification needs labels. In the Supplementary Material, we discuss how ExtraMAE and benchmarks generate labels for synthetic time series. We also defer the introduction to three datasets (i.e., Wafer, IPD, and Berry) to the Supplementary Material. 

\begin{table}[htb]
    \caption{Classification scores}
    \label{tab:class_scores}
      \begin{center}
        \begin{tabular}{llll}
            \toprule
            Method & Wafer & IPD & Berry \\
            \midrule
            ExtraMAE & $72.3    \pm 0.92$ & $83.9 \pm 4.05$ & $61.4 \pm 2.01$\\
            TimeGAN & $65.8\pm 1.01$ & $54.3 \pm 5.83$ & $44.5 \pm 1.71$\\
            RCGAN & $63.8 \pm 1.15$ & $57.0 \pm 5.58$ & $56.1 \pm 1.20$\\
            C-RNN-GAN & $65.1 \pm 1.47$ & $63.3 \pm 4.82$ & $55.7 \pm 1.82$\\
            \cmidrule(lr){1-4}
            Original & $72.3\pm1.59$ & $81.3\pm 4.18$ & $79.2\pm 1.89$\\
            \bottomrule
        \end{tabular} 
        \end{center}
\end{table}

\subsection{Imputation experiments}
Previous unsupervised models cannot manage missing data in the original time series. The supervised training, however, enables ExtraMAE to impute missing data. Imputation requires the model to capture the complex distribution and the temporal dependencies in multivariate time series. ExtraMAE efficiently learns the temporal dynamics by estimating extrapolation density. Therefore, we expect ExtraMAE scales well on imputation. We select a range of methods with different theoretic backups as the benchmarks for imputation. Table ~\ref{tab: imputation} compares the imputation results of ExtraMAE and benchmarks. We note that ExtraMAE beats all benchmarks with markedly significant improvements. Comparing with the current state-of-the-art Brits, ExtraMAE decrease $97\%$ mean squared error and $86\%$ mean absolute error on the Stock dataset ($97\%\approx\frac{|0.0007-0.0216|}{0.0216}$, $86\%\approx\frac{|0.0115-0.0802|}{0.0802}$). See the Supplementary Material for more details on imputation experiments. 

\begin{table}[ht]
\caption{Imputation results}
\label{tab: imputation}
\begin{center}
\begin{tabular}{lllll}
\toprule
Metric & Method & Stock & Sine & Energy \\
\midrule
             
             & Mean & $.0531\pm .0001$ & $.0540 \pm .0003$ & $.0347\pm .0000$\\
             & Median & $.0652\pm .0002$ & $.0586 \pm .0004$ & $.0353\pm .0000$\\
Mean squared & Soft & $.0215\pm .0547$ & $.0555 \pm .1291$ & $.0512 \pm .0861$\\
error (MSE)  & KNN & $.0214 \pm .0547$ & $.0496 \pm .1311$ & $.0434 \pm .0844$\\
             & Brits & $.0216 \pm .0039$ & $.1608 \pm .0060$ & $.0092 \pm .0054$\\
             & ExtraMAE & $.0007 \pm .0001$ & $.0069 \pm .0059$ & $.0088 \pm .0009$\\

\midrule
              
              & Mean & $.1844\pm .0002$ & $.1850 \pm .0005$ & $.1441 \pm .0000$\\
              & Median & $.1688\pm .0002$ & $.1801 \pm .0005$ & $.1416 \pm .0000$\\
Mean absolute & Soft & $.0533\pm .1046$ & $.1066 \pm .1755$ & $.1205 \pm .1496$\\
error (MAE)   & KNN & $.0500 \pm .1061$ & $.0716 \pm .1863$ & $.0920 \pm .1433$\\
              & Brits & $.0802 \pm .0156$ & $.0757\pm .0245$ & $.0420 \pm .0241$\\
              & ExtraMAE & $.0115 \pm .0032$ & $.0494 \pm .0214$ & $.0365 \pm .0073$\\

\bottomrule
\end{tabular}
\end{center}
\end{table}

\textbf{Mean.} The mean method \cite{little} replaces the missing entries with the mean of each column. 
    
\textbf{Median.} The median method \cite{little} replaces the missing entries with the median of each column. 

\textbf{Soft.} Soft \cite{soft} provides a regularised low-rank solution for large-scale matrix completion algorithms. It iteratively replaces the missing entries with the outcomes of soft-thresholded SVD \cite{svd}.
    
\textbf{KNN.} KNN \cite{knn} replaces the missing entries with a weighted combination of $k$ most related cases in the whole dataset. 
    
\textbf{Brits.} Brits \cite{brits} is the current state-of-the-art method in time series imputation. It directly learns the missing entries in a bidirectional recurrent dynamical system. 

\subsection{Main properties}
\textbf{Mask ratio} We first study the influence of mask ratios. We try ExtraMAE under different mask ratios and evaluate the practicality of corresponding synthetic data by prediction scores. The optimal mask ratio is $50.0\%$ for Energy, $16.7\%$ for Sine, and $4.2\%$ for Stock. Surprisingly, our ExtraMAE beats the runner-up even under extreme mask ratios ($91.7\%$ for Energy, $91.7\%$ for Sine, and $50.0\%$ for Stock). Table ~\ref{tab:extreme_high} shows the performance of ExtraMAE under optimal mask ratios and extreme high mask ratios, respectively. In the Supplementary Material, we train ExtraMAE with different combinations of mask size and number of masks. Experiments show that ExtraMAE generates practical yet diverse synthetic data under different mask ratios. We hypothesize that this behavior benefits from the latent representations and temporal dynamics ExtraMAE learned. In future work, we want to design one unified ExtraMAE to deal with the time series of different mask ratios. We also plan to better manage synthetic data by studying how to disentangle the latent space of ExtraMAE. Another exciting idea is adding controllable noises to manage the diversity of synthetic time series. 

\begin{table}[htb]
    \caption{Optimal mask ratio ExtraMAE \textit{vs.} high mask ratio ExtraMAE \textit{vs.} runner-ups. We evaluate the practicality by prediction scores. The lower, the better. }
    \label{tab:extreme_high}
        \begin{center}
        \begin{tabular}{lccc}
            \toprule
            Method & Stock & Sine & Energy \\
            \midrule
            Optimal ratio & $4.2\%$ & $16.7\%$ & $50.0\%$\\
            ExtraMAE   & $.037\pm .000$ & $.101 \pm .000$ & $.256 \pm .001$\\
            \cmidrule(lr){2-4}
            High ratio & $50.0\%$ & $91.7\%$ & $91.7\%$\\
            ExtraMAE   & $.051\pm .000$ & $.106 \pm .000$ & $.299 \pm .003$\\
            \cmidrule(lr){2-4}
            Runner-up  & $.053\pm .001$ & $.124 \pm .001$ & $.302 \pm .002$\\
            \bottomrule
        \end{tabular}
        \end{center}
\end{table}

\textbf{Ablation study} We then study alternative architectures for ExtraMAE. In the initial version of ExtraMAE, we take some classic ideas from the literature. \cite{initialize} suggest initializing model parameters by an autoencoder. \cite{timegan} proposes a supervised embedding loss to better capture temporal dynamics. We incorporate these modules into ExtraMAE and then ablate them one by one. Table ~\ref{tab: ablation} shows that both autoencoder pretraining and supervised embedding are redundant. Autoencoder pretraining and supervised embedding introduce mismatches between training and generation—this mismatch results in worse fidelity and practicality. Therefore, our final version ExtraMAE only keeps the reconstruction loss (See Figure ~\ref{fig: training}). We defer the ablation study on mask tokens to Supplementary Material.  

\begin{table}[htb]
    \caption{Ablation study on ExtraMAE. We evaluate practicality by prediction score (Pred) and fidelity by discrimination score (Disc). For both scores, the lower, the better. We denote autoencoder pretraining as A and use E to stand for supervised embedding loss. R is short for reconstruction loss. Therefore, ExtraMAE in AER mode is pre-trained as a vanilla autoencoder and then optimized by supervised embedding loss and reconstruction loss. Similarly, we denote the other three modes, ER, AR, and R. }
    \label{tab: ablation}
      \begin{center}
        \begin{tabular}{lll}
        \toprule
        Mode & Pred & Disc\\
        \midrule
        AER & $0.037939 \pm 0.000189$ & $0.084789 \pm 0.046658$ \\
        AR & $0.037846 \pm 0.000189$ & $0.085402 \pm 0.069082$ \\
        ER & $0.037195 \pm 0.000172$ & $0.132810 \pm 0.047714$ \\
        R & $0.037068 \pm 0.000150$ & $0.071965 \pm 0.020396$ \\
        \bottomrule
        \end{tabular}
        \end{center}
\end{table}

\section{Conclusions}
Simple algorithms that perform well lie at the core of deep learning. This study shows that masked autoencoder with extrapolator (ExtraMAE) is light but superb self-supervised model for time series generation. ExtraMAE proposes an extrapolator to learn extrapolation density in the latent space. The synthetic time series generated by ExtraMAE demonstrates excellent practicality and fidelity. Our approach is markedly better than all benchmarks while smaller and faster in a minimalist design. It also scales well on downstream tasks such as prediction, classification, and imputation. To our surprise, ExtraMAE infers complex original signals even under an extreme mask ratio near $92\%$. We hope this property will inspire future work. Time series generation may now usher in the era of supervised learning. 

\bibliographystyle{unsrt}  

\bibliography{references}

\end{document}